\documentclass[11pt,letterpaper]{article}

\usepackage{siunitx,etoolbox}
\usepackage{tabularx}
\usepackage{booktabs}
\usepackage{caption}
\usepackage[usenames, dvipsnames]{color}
\usepackage{subcaption}
\usepackage{setspace}
\usepackage{hyperref}
\usepackage{multirow}
\usepackage{mdframed}
\usepackage{authblk}
\usepackage{pslatex}
\usepackage{graphicx}
\usepackage[margin=1.0in]{geometry}
\usepackage{pgfplots}
\usepackage{xcolor, colortbl}
\hypersetup{
    colorlinks,
    linkcolor={cyan!20!red},
    urlcolor={cyan!80!green},
    citecolor={cyan!80!blue}
}
\definecolor{light-gray}{HTML}{F0F0F0}

\usepackage{gb4e}  
\noautomath

\usepackage{amsmath}

\title{A neural network to classify metaphorical violence on
cable news}

\author{Matthew A.~Turner\thanks{\href{mailto:mturner8@ucmerced.edu}{mturner8@ucmerced.edu}}}

\affil{\footnotesize Cognitive and Information Sciences, University of California, Merced, Merced, CA, USA}

\date{}

\begin{document}

\maketitle

\begin{abstract}
I present here an experimental system for identifying and annotating metaphor 
in corpora. It is designed to plug in to 
  Metacorps,
an experimental web app for annotating metaphor. As Metacorps users annotate metaphors,
the system will use user annotations as training data. When the system is 
confident, it will suggest an identification and an annotation. Once approved by the user, this becomes more training data. This naturally allows for transfer learning, where the system can, with some known degree of reliability, classify one class of metaphor after only being trained on another class of metaphor. For example, in our metaphorical violence project, metaphors may be classified by the network they were observed on, the grammatical subject or object of the violence metaphor, or the violent word used (hit, attack, beat, etc.).
\end{abstract}

\section{Introduction}\label{introduction}

Metaphor is often thought of as a decorative instrument of literature.
It is more scientifically productive to understand metaphor a basic
human cognitive capacity used to represent relationships between concepts.
Mathematics, for example, may be understood as an elaborate metaphorical
scaffolding where ``embodied'' concepts are everywhere, such as the fact
that multiplying by \(e^{i\theta}\) rotates a point in the complex plane by \(\theta\)
radians. Rotation is something we do every time we drive a car, and in many 
other situations, so in
that way \(e^{i\theta}\) is physically intuitive because we have been
rotating different physical objects our entire 
lives \cite{Nunez1999, Núñez2000, Alibali2014}. 
In this paper we present a system to identify metaphorical violence (MV) 
on cable news surrounding presidential elections\footnote{Code publically available: \url{https://github.com/mt-digital/metacorps-nn}}. Just as our intuitions about
physical transformations
of physical objects help us reason about abstract mathematics, our intuitions
about violence serve as heuristics for rather more complex political events 
(for a good discussion of the
importance of heuristics to cognition, see \cite{Gigerenzer2009}).
Specifically, we present a neural network classifier for
determining whether or not a phrase is MV or not. This effort is an
important part of a larger effort to improve the throughput of an
existing metaphor annotation software application we are developing
called \emph{Metacorps}\footnote{\url{https://github.com/mt-digital/metacorps}}. 
Knowing what metaphors are said and
when in a society is important because metaphors are representations of
a society's conceptual relationships. Large-scale observations of
metaphor use are currently limited because it is a time-intensive task
for humans to do. We demonstrate that even with a modest gold-standard
dataset, a neural network system can distinguish, to around 85\%
accuracy, MV from non-MV. After introducing the motivation, we present
the data, our methods, and analyze the performance of some candidate
neural network classifiers. We close with a discussion of the promise
and challenges of integrating this system with teams of human
annotaters.

Evidence is growing that choice of metaphor, or whether or not to use
metaphor, has real consequences on cognition and behavior in general
\cite{Lakoff2014} and for politics \cite{Matlock2012}. 
A recent behavioral experiment showed more specifically that when
trait-aggressive people, who are more aggressive independent of context,
are exposed to metaphorical violence in political speeches, their support for
real violence against politicians increases.
Metaphor use is revealing:
spoken metaphors are either representative of the speaker's conceptual
system, or they are intended to cause the hearer to activate the metaphor's
conceptual links, or both.
So, if you want people to take climate change more seriously, you would
do better to frame it as a ``war'' instead of a ``race'' \cite{Flusberg2017}.
While the previous claim is supported by data, context changes the production
and effects of metaphor use. If we can use machine learning to identify and
annotate metaphor, we will understand much better than we do now just when
and why one metaphor is chosen over another.

One shortcoming in behavioral studies of metaphor is that
cultural context is not part of the experiment---in other words, the experiment
lack an important element of ecological validity. Linguists 
are increasingly looking to cultural context as
an important factor for explaining metaphor use \cite{Kovecses2010}. 
Another criticism
says we should not be asking ``if'' choice of metaphor influences
reasoning, but ``when'' does a particular metaphor have a causal effect
on reasoning \cite{Steen2014}? The contribution presented here helps solve
this contextual shortcoming, in that our data is timeseries data, so
current events such as the presidential debates are happening in the
cultural background. Further context is ideological context, with cable
news channel as proxy for ideology \cite{Pew2014, King2017}. More
generally, the context is English-speaking American cable television news.
Our analysis enables us to observe how metaphorical violence varies between 
cable news networks.

An example metaphorical phrase might be the headline ``Bernie attacks
Clinton Foundation in first debate.'' On the contrary, ``Terrorist attack kills
US Ambassador to Libya'' is clearly not a metaphor. Metaphor serves a pragmatic
purpose: it is a much shorter, if incomplete, version of what might be
the non-metaphorical way to explain that actually ``Bernie claimed that
the Clinton Foundation had wrongly accepted funds.'' In a concurrent
project, we have developed and fit a simple statistical model to the
frequency of metaphorical violence (MV) use, and found that MV usage
increases in frequency for all networks, but with different timing in
2012 and 2016 \cite{Turner2018a}.

While we are able to get novel and interesting results with
human-generated annotations, machine learning could help us increase our
signal power. For the study just mentioned, in 2012 and 2016 we had to
limit our corpus to include only the top two most-watched shows on each
of the three networks. As mentioned, we limited ourselves there to three
violent words. This is because manual annotation is a time-consuming
process. Often, relevant parts of the episode must be watched to finish
an annotation, which currently requires navigation of the TVNA website.
The first use of Metacorps was to increase the efficacy of human coders
by streamlining the annotation process. But by building our own
annotation web application and data model, we can incorporate a new
service whose prototype is introduced in this report: a neural network
that classifies potential metaphorical violence as metaphor or not. The
next section details the machine learning task, describes the gold
standard dataset and the training/validation/test datasets.

\section{Data}\label{data}

Metacorps provides a data model and web interface for annotating
metaphor in a corpus. In its present state, all corpora are
pre-processed subcorpora of the closed captions hosted and curated by
the TV News Archive. The TV News archive
\url{http://archive.org/tv/details} provides video, audio, closed
captioning, and rich metadata for millions of hours of television news
from cable news channels, studied here, and local news. They provide an
HTTP API that provides JSON search results and episode metadata. To be
specific, \emph{episode} refers to a single showing of a particular
show, like \emph{The O'Reilly Factor}. Reruns are independent episodes
and must be excluded. A show's metadata includes links to its data,
which may be in the form of video, audio, or closed captions. This study
uses the closed captions. To programatically acquire data and build
transcripts from closed captionings we used Python software
\texttt{iatv} available on GitHub (\url{http://github.com/mtpain/iatv}).

Gold standard annotations were created using the Metacorps Python
software package and web application. Trained annotators
use the web application to indicate which phrases of potential metaphor
actually are metaphor, then fill in more information, such as who the
subject and object are of the metaphorical violence. Potential instances
of metaphorical violence are found by searching for key violent words;
in this study those words are \emph{hit}, \emph{beat}, and
\emph{attack}. The corpus we searched was of transcripts from the top
two shows on each of the three cable networks MSNBC, CNN, and Fox News from
the months September to November, 2012. 

\begin{figure*}[t]
  \caption{Schematic of Metacorps: the annotation tool-set that connects 
    cable news transcripts to annotators, annotations to meaningful data tables, 
    and soon annotations and analyses to our neural network classifier.}
  \centering
  \includegraphics[width=.9\textwidth]{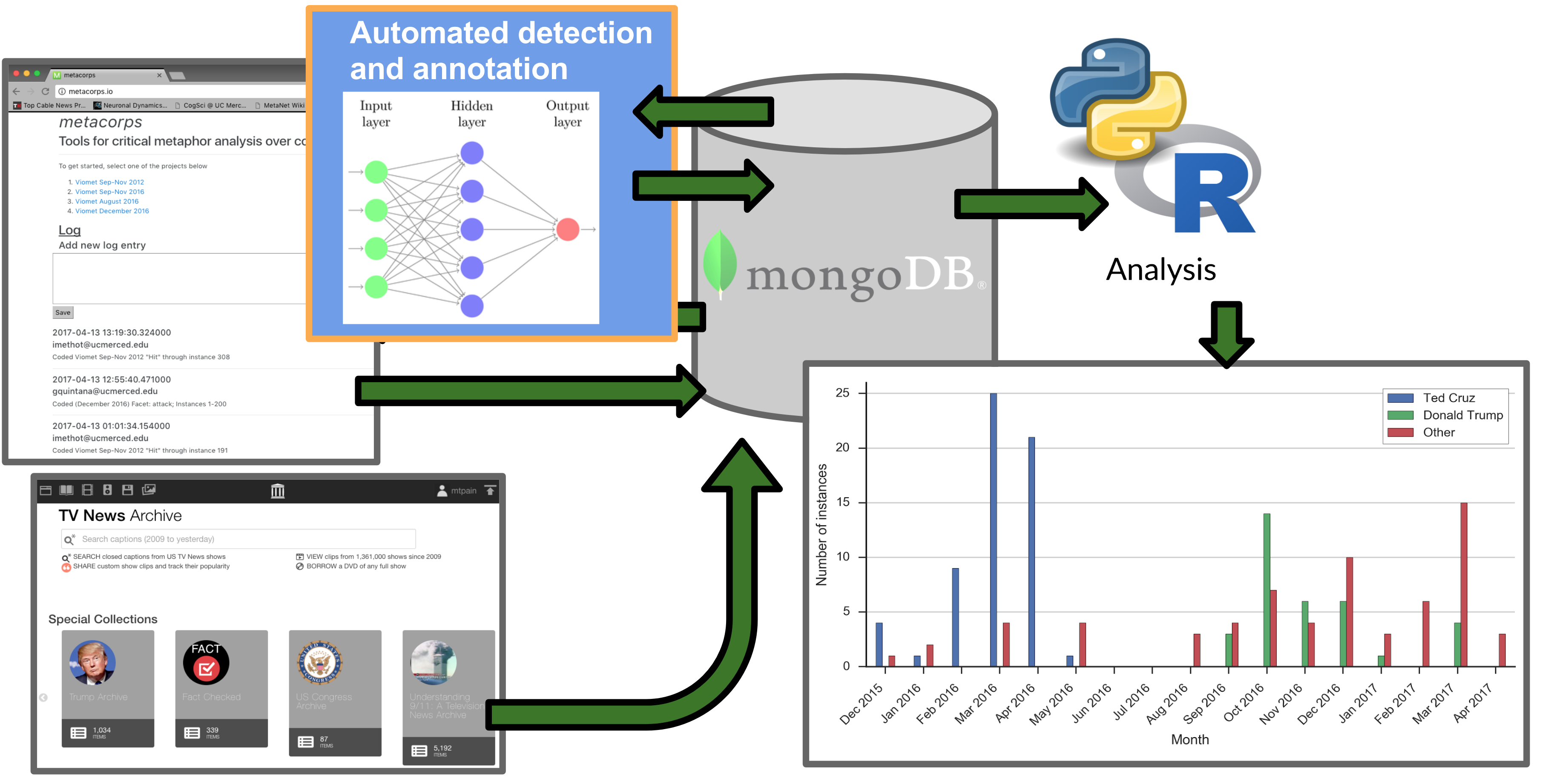}
\label{fig:metacorps}
\end{figure*}

To build the corpus, we used \textit{iatv} to download the transcripts from
the desired shows in the desired timeframe. \textit{Metacorps} provides 
a data model built on MongoDB and its object-document mapper mongoengine for
Python. Transcripts and other metadata are persisted. Other elements in the
data model provide fields to annotate these base transcripts, which are
linked. Human annotaters have suggested and agreed upon whether or not a
phrase is metaphorical violence. See Figure \ref{fig:metacorps} for a
sketch of the Metacorps data flow and where our new classifier fits in.
Metacorps also provides a data exporter that
created the gold standard dataset 
\href{http://metacorps.io/static/data/viomet-2012.csv}{viomet-2012.csv} used
to train, validate, and test our model. This dataset has 2538 rows, 791 of 
which were classified as metaphor, a prevalence of about 31\%. This dataset
first was split into 80\% pre-training and 20\% test datasets. The test dataset
was left alone, but the pre-training set was split once more: 80\% of the
pre-training rows became training and the rest validation rows. To create a 
balanced training set, metaphorical rows were resampled with replacement and
added to the training set until there were an equal number of metaphor and
non-metaphor rows. 

\section{Methods}\label{methods}

To accomplish this MV classification task, we experimented with deep
feed forward neural networks, encouraged by success on a similar task reported in
\cite{DoDinh2016}. Each hidden layer had 500 nodes, and we
tested 1-, 2-, 4-, and 6-hidden-layer models. To regularize we use
dropout and early stopping. Stochastic gradient descent with momentum
minimized the cross entropy loss. The inputs were vectors with 3300
elements: eleven words represented by their word2vec embeddings, which
come from the Google News pre-trained word2vec 
(\href{https://github.com/mmihaltz/word2vec-GoogleNews-vectors/raw/master/GoogleNews-vectors-negative300.bin.gz}{download}).
Each word embedding vector is concatenated with the next to create the
network input vector. We used the free and open source Python package 
\texttt{gensim} for loading the binary-formatted word2vec 
\cite{Mikolov2013, Mikolov2013a} word embeddings \cite{Rehurek2010}.
Neural network construction, training, and testing was done primarily in
\href{https://www.tensorflow.org/}{TensorFlow} 
\cite{GoogleResearch2015, Abadi2016a}. Performance 
analysis was done using Scikit-Learn \cite{Pedregosa2011}, 
and data table reading, writing, and 
subsetting was done with Pandas \cite{McKinney2013}. 
More details can be found by examining the
README and code for 
\href{https://github.com/mtpain/math292-fall17-project}{this project on GitHub}.

To investigate the performance of the different hyperparameterizations over
the number of layers and the learning rate, we executed twenty trials for
every combination of the number of layers (1, 2, or 4) and learning rate
(0.01 or 0.1). To do this we utilized the MERCED computing cluster hosted
at UC Merced. We then calculated the average precision, sensitivity, specificity,
and AUC for each of these; the results are described in the next section and
summarized in Table~\ref{tab:main}.

\section{Results}\label{results}

Looking to Table~\ref{tab:main}, we see promising results for a first attempt
at using a neural network for classification with our cable news corpus. 
The maximum AUC reached was 0.92, with a specificity of .926, for 
the four-layer neural network using a learning rate of 0.01. While specificity
is rather good here, the precision and sensitivity are relatively lower.
This is saying it is easier for the system to judge a 
true negative than a true positive. This is what we might expect given that we 
have more negative examples than positive examples. 

\begin{table}[h!]
  \caption{
    The best performing model on three of the four measures we use here
    is four layers with a learning rate of 0.01, shown in bold. 
    Only its sensitivity is 
    outperformed by the single-layer trained with a 0.01 learning rate.
    A learning rate of 0.5 and six layers were also tried, 
    but these often failed to converge, so they were excluded from this 
    analysis. Average of five trials.
  }
  \centering
  \begin{tabular}{rrcccc}
  \toprule
    &      &  sensitivity &  specificity &  precision &    auc \\
  N layers & learning rate &              &              &            &        \\
  \midrule
  1 & 0.01 &        0.796 &        0.890 &      0.768 &  0.918 \\
    & 0.10 &        0.632 &        0.918 &      0.780 &  0.886 \\
  2 & 0.01 &        0.764 &        0.894 &      0.766 &  0.914 \\
    & 0.10 &        0.758 &        0.886 &      0.740 &  0.910 \\
    \textbf{4} & \textbf{0.01} & \textbf{0.702} & \textbf{0.926} &      \textbf{0.816} &  \textbf{0.920} \\
    & 0.10 &        0.618 &        0.902 &      0.720 &  0.882 \\
  \bottomrule
  \end{tabular}
  \label{tab:main}
\end{table}

\section{Discussion}\label{discussion}

At this point, we need to know much more than we do about what the system is
doing. An immediate result we should obtain is model performance for subsets
of the test set taken by violent word, and by network, e.g. answer the
question ``How well does the system predict metaphorical violence when the
word \emph{attack} is used compared to \emph{hit} or \emph{beat}?'' What about
for individual cable news networks? Can the system more accurately classify
MV on MSNBC than Fox News?
Clearly more work needs to be done to select optimal hyperparameters, as we
have only shown two. We also need more than five trials to understand model
performance.

It would be interesting to try different architectures. Long-short term
memory models are often used for natural language processing applications,
as in neural machine translation \cite{Wu2013, Sutskever2014}. However, the
structure of the inputs is really a 2D structure, like an image. So convolutional
neural networks may detect correspondences between co-occurring features of
words that other network architectures would not. An architecture for detecting
whether a pair of words are used metaphorically, called ``supervised similarity
networks,'' was introduced in \cite{Rei2017}. While pair detection like this
is not directly applicable for our task, the results are promising, and 
this approach should be investigated more. 

Also in \cite{Rei2017}, the authors explore using ``cognitive'' embeddings.
There are many embeddings availalbe, so an interesting and important 
empirical study would be to test how well different word embeddings perform
for this task. Perhaps word embeddings built from the corpus itself would
outperform word embeddings from the Google News word2vec model, if we could
produce quality embeddings---not necessarily an easy task.

While there are still many details to fill in, deep neural networks using 
TensorFlow and the Google News word embeddings seem to provide the capability
to solve a real-world problem: that of discerning metaphorical statements
from non-metaphorical statements. It is a demonstration of the power of 
word embeddings, and the representational capabilities of neural networks. 
While some put great effort into building detection systems based on
\textit{a priori} theories of how metaphor works \cite{Dodge2015}, 
neural networks offer us
the possibility not just of automated detection, but perhaps can even guide
cognitive scientists towards a deeper understanding of how metaphorical 
meaning-making works in the brain. If nothing else, this system can already
improve the throughput of metaphor annotation systems by suggesting a label 
that will be correct with high likelihood. At first human annotators will
have to deal with correcting mistakes, but if many mouse clicks can be saved
just the same, that will be a help.

\bibliographystyle{unsrt}
\bibliography{/Users/mt/workspace/papers/library.bib}

\end{document}